\documentclass[10pt,twocolumn,letterpaper]{article}
\usepackage[table]{xcolor}
\usepackage{cvpr}              % To produce the CAMERA-READY version
\usepackage[accsupp]{axessibility}  % Improves PDF readability for those with disabilities.

\definecolor{cvprblue}{rgb}{0.21,0.49,0.74}
\usepackage[pagebackref,breaklinks,colorlinks,citecolor=cvprblue]{hyperref}
\usepackage{paralist}
\usepackage{indentfirst}
\usepackage{multirow}
\usepackage{makecell}
\def\paperID{2665} % *** Enter the Paper ID here
\def\confName{CVPR}
\def\confYear{2024}

\title{Task-Driven Exploration: Decoupling and Inter-Task Feedback for Joint Moment Retrieval and Highlight Detection}
\newcommand\blfootnote[1]{%
  \begingroup
  \renewcommand\thefootnote{}\footnote{#1}%
  \addtocounter{footnote}{-1}%
  \endgroup
}
\author{
Jin Yang, Ping Wei$^{\ast}$, Huan Li, Ziyang Ren\\
National Key Laboratory of Human-Machine Hybrid
Augmented Intelligence\\
Institute of Artificial Intelligence and Robotics, Xi'an Jiaotong University\\
{\tt\small \{jin.yang, lh875056558, rzyrzy\}@stu.xjtu.edu.cn, pingwei@xjtu.edu.cn}
}

\begin{document}
\maketitle
\begin{abstract}
Video moment retrieval and highlight detection are two highly valuable tasks in video understanding, but until recently they have been jointly studied. Although existing studies have made impressive advancement recently, they predominantly follow the data-driven bottom-up paradigm. Such paradigm overlooks task-specific and inter-task effects, resulting in poor model performance. In this paper, we propose a novel task-driven top-down framework TaskWeave for joint moment retrieval and highlight detection. The framework introduces a task-decoupled unit to capture task-specific and common representations. To investigate the interplay between the two tasks, we propose an inter-task feedback mechanism, which transforms the results of one task as guiding masks to assist the other task. Different from existing methods, we present a task-dependent joint loss function to optimize the model. Comprehensive experiments and in-depth ablation studies on QVHighlights, TVSum, and Charades-STA datasets corroborate the effectiveness and flexibility of the proposed framework. Codes are available at \href{https://github.com/EdenGabriel/TaskWeave}{github.com/EdenGabriel/TaskWeave}.
\blfootnote{$^*$ Corresponding author} 

% \blfootnote{
% $^\star$ Corresponding author
% }

\end{abstract}

% resulting in the lack of interpretability in the model.     
\section{Introduction}
\label{sec:intro}

\begin{figure}[tbp]
  \centering
  \includegraphics[width=1\linewidth]{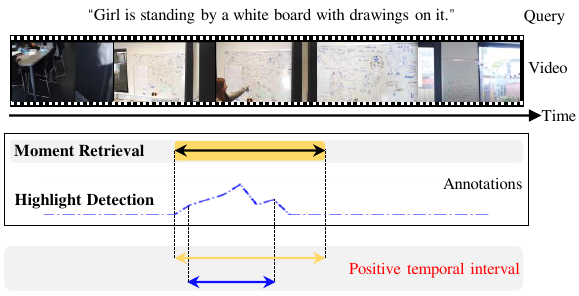}
  \caption{Although the positive temporal intervals of moment retrieval and highlight detection exhibit high overlap, they pursue different objectives. 
  % Given the video and a query requested by the user, TaskWeave aims to simultaneously retrieve relevant moments and detect the highlightness of the video. It also in-depth investigate the influence between these tasks.
  % Illustration of different paradigms for addressing joint MR and HD tasks. (a) Data-driven: the paradigm adopted by existing methods, these methods focus on fusing multi-modal data to capture common features for accomplishing joint tasks. (b) Task-driven: the novel paradigm we propose, which operates from the task-centric perspective, achieving joint tasks through task decoupling and inter-task feedback.
  }
  \label{fig:introduction}
\end{figure}

As videos are prevailing in a wide range of applications, the diversity and massive scales of video content have posed unprecedented challenges in finding relevant moments based on user queries. To this end, the moment retrieval (MR) \cite{zhang2019man,zhang2020learning,zhang2023tsgsurvey} and highlight detection (HD) \cite{yao2016highlight,badamdorj2022contrastive} tasks have emerged recently. MR aims to retrieve video moments that are relevant to the given query \cite{gao2017tall,10075491tsg}. HD aims to predict the clip-level saliency scores in the video \cite{sun2014ranking}.
% This is largely attributed to the richness, diversity, and engaging nature of its content. 

As MR and HD tasks are closely related, they have been jointly addressed and achieved breakthroughs recently \cite{lei2021moment-detr, liu2022umt,moon2023qd-detr,lin2023univtg,jang2023knowing,xu2023mh}. The existing joint approaches in general utilize a shared backbone to learn the multi-modal features as the common representations for MR and HD. Then a MR prediction head is employed for moments localization and a HD prediction head predicts saliency scores. These methods adhere to the bottom-up, data-driven paradigm, i.e. they capture common features from the input data and then utilize the features for different tasks. The effectiveness of these methods is built upon the premise of the high correlation between MR and HD. As illustrated in \cref{fig:introduction}, the task MR and HD share the identical query and video inputs, additionally exhibiting a substantial temporal overlap between their respective positive temporal intervals. 

% primary bottom-up
However, the bottom-up, data-driven paradigm tends to excessively rely on the common features, but overlooks the inherent specific characteristics of MR and HD. This tendency might simplify the joint modeling as a problem of feature fusion, without considering the interplay between the two tasks. The distinct objectives pursued by MR and HD rely on distinct task-specific characteristics. Unfortunately, existing methods overlook the specificity.

 %We believe that existing methods adhere to the bottom-up, data-driven paradigm. While this paradigm is straightforward and easy to implement, it tends to excessively rely on fundamental multi-modal features, lacking a comprehensive understanding of the inherent task characteristics. This tendency might simplify the tasks as problems of feature fusion, overlooking the potential interplay between both tasks. In addition, in \cref{fig:introduction}, the distinct objectives pursued by MR and HD contribute to their respective specificities. Unfortunately, all existing methods have overlooked these specificities.%

To address the aforementioned issues, we believe it is essential to leverage the fundamental multi-modal data to mine commonalities across tasks (bottom-up), while also strengthen the awareness of task-specific characteristics (top-down). To this end, we propose a novel paradigm TaskWeave from a task-driven perspective. The key idea is to jointly address the tasks MR and HD by considering the commonality, specificity, and interplay of MR and HD.

To effectively capture the commonality and specificity, we design a task-decoupled unit, i.e. a shared expert to capture common features and two task-specific experts to acquire the distinct characteristics. In order to investigate the interplay between MR and HD in-depth, we design an inter-task feedback mechanism. It converts the predictions of MR/HD into mask information, which are fed back to the input of the HD/MR prediction head. Furthermore, we introduce a principled task-dependent joint loss in which the task-specific weights are dynamically adjusted, rather than manually tuned.
 
We conduct experiments on the QVHighlights \cite{lei2021moment-detr} dataset to validate the effectiveness of the proposed method. Moreover, we also conduct experiments for two individual tasks on the datasets Charades-STA \cite{gao2017tall} (moment retrieval) and TVSUM \cite{song2015tvsum} (highlight detection). The proposed approach outperforms the existing methods. 

The key contributions of this paper are three folds.
\begin{compactitem}
	\item [1.] It proposes a novel task-driven, top-down framework for joint moment retrieval and highlight detection.
	\item [2.] It introduces a task-decoupled unit, an inter-task feedback mechanism, and a principled task-dependent loss.
	\item [3.] It achieves state-of-the-art performance on three datasets. The ablation study validates the methods.
\end{compactitem}

% The common features and task-specific features are then fused for prediction. 
\section{Related Work}
\label{sec:relatedwork}

\begin{figure*}[tbp]
  \centering
  \includegraphics[width=1\linewidth]{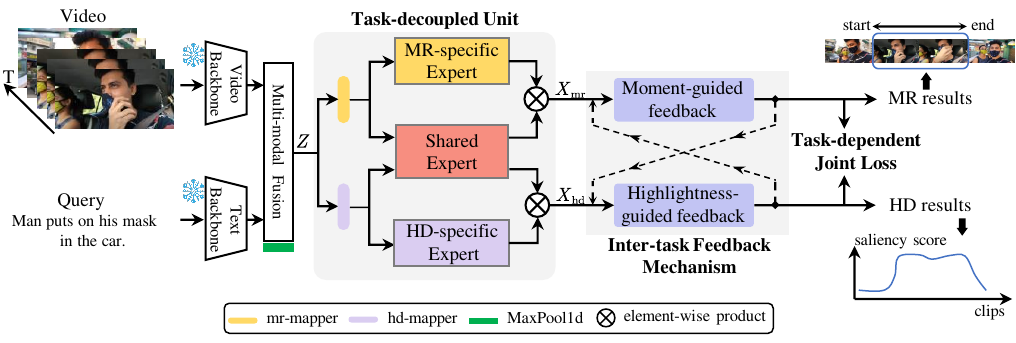}
  \caption{The overall pipeline of the proposed task-driven model TaskWeave. We propose the task-decoupled unit to capture task-specific and common features. Various experts can adopt different network implementations, showcasing the flexibility of the model. Inter-task feedback mechanism is designed to investigate the influence between both tasks. There are two feedback manners: Moment-guided and Highlightness-guided feedback. The principled task-dependent joint loss is introduced for jointly optimize the model.
  }
  \label{fig:pipeline}
\end{figure*}

We review the related work from four aspects.

\textbf{Moment Retrieval.} Existing approaches for MR mainly include two groups. One group follows a two-stage procedure \cite{anne2017localizing,gao2017tall,zhang2019man,zhang2020learning}, which involves generating candidate temporal intervals and ranking them based on the correlation with the query. The other group directly regresses the temporal interval based on the aligned visual-text features \cite{yuan2019find,chen2019localizing,zhang2020span,mun2020local,zeng2020dense}. Moreover, most datasets provide only one moment annotation for each video-query pair \cite{gao2017tall,anne2017localizing}, which does not align with real-world scenarios.

% 篇幅所限，先注释
% Furthermore, some studies have pointed out significant temporal bias issues \cite{zhou2021embracing,lei2020tvr} in previous benchmark datasets. Specifically, moments related to queries tend to appear more frequently at the beginning of videos than at the end \cite{zhou2021embracing}. Additionally, the previous benchmark datasets provided only one moment annotation for each video-query pair \cite{gao2017tall,anne2017localizing}, which does not align with real-world scenarios.

\textbf{Highlight Detection.} The saliency score in the highlight detection represents the relevance of a video clip to the given query. Most prior highlight detection benchmark datasets are query-agnostic \cite{sun2014ranking,garcia2018phd}. The saliency scores for video clips remained constant regardless of the query. As a result, some previous approaches treated HD as a solely visual task \cite{sun2014ranking,yao2016highlight,wei2022learning,badamdorj2022contrastive}.
% 篇幅所限，先注释
% Highlight detection aids in creating more concise video summaries and facilitating quick content comprehension, thus endowing it with significant practical utility.

MR and HD tasks are traditionally studied separately. They have been jointly addressed recently as the introduction of QVHighlights dataset \cite{lei2021moment-detr}. QVHighlights provides multiple moment-annotations for each query and ensures these moments are uniformly distributed throughout the video. It also provides query-dependent highlightness annotations. With QVHighlights, the model Moment-DETR \cite{lei2021moment-detr} is proposed for joint MR and HD. Following Moment-DETR, a growing number of approaches have been proposed to accomplish the joint task \cite{liu2022umt,moon2023qd-detr,jang2023knowing,lin2023univtg}. However, these methods adopt the data-driven and bottom-up paradigm. Different from them, we propose a novel task-driven and top-down paradigm.
% 篇幅所限，先注释
% Although MR and HD tasks are closely related, they are traditionally studied separately. This phenomenon has been changed with the introduction of the QVHighlights dataset \cite{lei2021moment-detr}. It is the first dataset that contains both MR and HD tasks. 
% It also provides query-dependent highlightness annotations. 
% In order to overcome the limitations of previous MR and HD benchmark datasets, QVHighlights provides multiple moment-annotations for each query and ensures these moments are uniformly distributed throughout the video.

\textbf{Vision Transformers.} Transformer-based models \cite{attention2017} have brought huge achievement in both the image and video related domains \cite{carion2020detr,dosovitskiy2020vit,yang2023cross,yang2023gated,li2023inverse,li2022asymmetric}. One of the most well-known methods is DETR \cite{carion2020detr}, which regards object detection as a set prediction problem. Its end-to-end prediction procedure eliminates the intermediate or post-processing steps. On the other hand, some studies have employed cross-attention mechanism \cite{moon2023qd-detr,wei2020multi,mercea2022audio} to inject multi-modal data into the Transformer architecture. In this paper, we also adopt a DETR-like architecture \cite{li2022dn,liu2022dab}. However, different from those methods, we incorporate diverse network architectures for feature extraction. Our focus lies in the collaboration of the network architectures.

\textbf{Multi-task Learning.} Multi-task learning (MTL) aims to train a single model for multiple tasks \cite{caruana1997multitask}. The most straightforward approach is to utilize a shared backbone to extract common features, which relies on the data-driven manner. However, it often leads to suboptimal performance for unrelated tasks and lacks flexibility. In response, some studies introduce MOE \cite{jacobs1991moe} and MMOE \cite{ma2018mmoe}, which utilize a set of the shared experts in place of the shared bottom layer. To mitigate the seesaw phenomenon of multi-task learning, PLE \cite{tang2020ple} employs separate experts for each task while still retaining the shared experts. From the perspective of MTL, existing approaches for joint MR and HD follow the shared bottom paradigm. Moreover, directly applying MTL methods to joint MR and HD does not yield favorable results. Existing MTL methods are complex, therefore direct usage of them leads to a sharp increase in model complexity. This might decrease the performance. To this end, we aim to design an effective task-driven framework to jointly address MR and HD.

% 篇幅所限，先注释
% In MOE \cite{jacobs1991moe}, all tasks share a gating network, whereas MMOE \cite{ma2018mmoe} optimizes MOE by providing the gating network for each task. 
\section{Methodology}
\label{sec:methodology}

\subsection{Overview}
% Finding moments in the video that are relevant to the given text query is the common objective of both moment retrieval and highlight detection. 

% \textbf{Problem Statement.} Given a text query with $W$ words and an untrimmed video composed of $N$ clips, the objective of the joint moment retrieval and highlight detection is to localize the center coordinate $q_c$ and width $q_w$ of temporal intervals within the video, in addition to ranking clip-wise saliency scores $\left\{s_1,s_2,...,s_N\right\}$.
Given a text query with $W$ words and an untrimmed video composed of $N$ clips, the objective of the joint moment retrieval (MR) and highlight detection (HD) is to localize the center coordinate $q_c$ and width $q_w$ of temporal intervals that are relevant to the text query, in addition to ranking clip-wise saliency scores.

\textbf{Architecture Overview.} Given the high correlation between moment retrieval and highlight detection tasks, the most intuitive approach is using a shared backbone in conjunction with two task-specific prediction heads. It is a data-driven paradigm that is commonly employed by previous methods, for its simplicity and ease of implementation. However, the task-specific characteristics are inherently present, since MR and HD have distinct objectives. Additionally, the interplay between MR and HD should also be considered, which can enhance the model's performance.

The overall pipeline of our approach is illustrated in \cref{fig:pipeline}. We employed the frozen video/text-encoder backbones to extract the video/text features, while ensuring their dimensions remained consistent at $D$ by the projection. These features are utilized for multi-modal fusion through methods such as cross-attention or concatenation, resulting in query-related video representations $Z \in \mathbb{R} ^{N\times D}$. In our paper, we extract these representations through the cross-attention. After cross-attention, inspired by \cite{tang2023temporalmaxer,shi2023tridet}, a 1D Max Pooling (MaxPool1d) with the kernel size 5, stride 1 and padding 2 is utilized to eliminate the rank loss problem in the attention mechanism. These representations are fed into the task-decoupled unit to capture the task-related features $X_{\mathrm{mr}} \in \mathbb{R} ^{N\times D}$, $X_{\mathrm{hd}} \in \mathbb{R} ^{N\times D}$. Then, we employ the task-specific decoders with inter-task feedback mechanism to make predictions for moments localization and clip-wise saliency scores. We introduce the principled task-dependent loss to jointly optimize the model.

% Latex书写规范：section/subsection之间空两行 段落/图/表之间空一行
% 阐述component的思路：先简要阐述motivation，然后阐述具体方法，阐述具体设计时可以描述component的优势
\subsection{Task-decoupled Unit}
Since moment retrieval and highlight detection have distinct objectives, the specificity of each task should be considered, rather than solely focusing on their commonalities. For this purpose, we propose a task-decoupled unit from a task-driven perspective to capture the task-related features, which involves task-specific features and common features.

The task-decoupled unit is depicted in \cref{fig:pipeline}. Inspired by the attention mechanism \cite{attention2017}, the query-related video representation $Z$ is initially fed into two task-specific mappers, mr-mapper and hd-mapper. Each mapper is implemented using one-layer feed-forward network. The output of each task-specific mapper is directed into both the respective task-specific expert network (MR-specific Expert or HD-specific Expert) and the Shared Expert network. 

Thanks to the design of our task-decoupled framework, each expert can employ various networks, such as convolutional network, Transformer, and feed-forward network. It offers more flexibility in addressing the joint MR and HD tasks. In our study, we also investigate the influence of different expert networks, referring to \cref{sec:experiments} for more details. Consequently, we can capture the task-related features $X_{\mathrm{mr}}$ and $X_{\mathrm{hd}}$ via element-wise product between the output of the task-specific expert and that of the shared expert. 

The MR-specific feature $X_{\mathrm{mr}}$ can be computed as:
\begin{equation}
\label{eq:task-decoupled unit}
    X_{\mathrm{mr}}=\mathcal{P} _{\mathrm{mr}}\left( \mathcal{M} _{\mathrm{mr}}\left( Z \right) \right) \otimes \mathcal{S} \left( \mathcal{M} _{\mathrm{mr}}\left( Z \right) \right),
\end{equation}
where $\mathcal{M} _{\mathrm{mr}}\left( \cdot \right)$ refers to the mr-mapper operation. $\mathcal{S}\left( \cdot \right)$ is the shared expert calculation. $\otimes$ denotes the element-wise product. $\mathcal{P} _{\mathrm{mr}}\left( \cdot \right)$ is the mr-specific expert calculation. The calculation method for the HD-specific feature $X_{\mathrm{hd}}$ follows the similar approach.

\begin{figure}[tbp]
  \centering
  \includegraphics[width=1\linewidth]{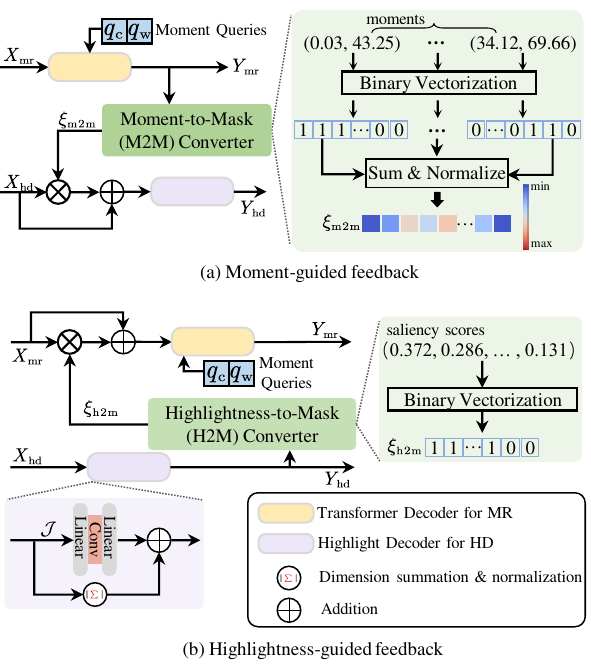}
  \caption{ Illustration of the inter-task feedback mechanism. (a) moment-guided feedback manner. (b) highlightness-guided feedback manner.
  }
  \label{fig:feedbacks}
\end{figure}

\subsection{Inter-task Feedback Mechanism}
It is imperative to investigate the interplay between the moment retrieval and highlight detection tasks for their significant correlation. However, existing methods overlook this aspect and employ two decoders for direct prediction.

To resolve this issue, we propose an inter-task feedback mechanism, which contains two task-specific decoders (Transformer decoder for MR and lightweight decoder for HD) and two feedback manners (moment-guided feedback and highlightness-guided feedback). The output of one task is transformed into the mask to assist another task in the feedback mechanism. All components of the inter-task feedback mechanism is detailed as follows.

\textbf{Transformer Decoder for MR.} Previous studies have demonstrated that it is effective to employ dynamic anchors as queries for the decoder within the DETR-like structure \cite{li2022dn,liu2022dab}. They iteratively update anchor boxes, thereby transforming the procedure for updating queries into the cascaded soft ROI-pooling. In our work, the Transformer decoder for MR follows the work \cite{liu2022dab}. Each moment query is represented by its temporal center coordinate $q_c$ and width $q_w$. The output of the Transformer decoder for MR is denoted as $Y_\mathrm{mr} \in \mathbb{R} ^{N_q\times 2}$. $N_q$ is the number of moment queries and set to 10 in our work. 

\textbf{Lightweight Decoder for HD.} The most straightforward method for the HD decoder would be utilizing one or more fully-connected layers, as seen in prior studies such as Moment-DETR \cite{lei2021moment-detr} and UMT \cite{liu2022umt}. However, such design overlooks the diversity of video-query pairs \cite{moon2023qd-detr}, which provides identical criterias for the saliency score prediction of each video-query pair. Although QD-DETR \cite{moon2023qd-detr} proposes a global saliency token to predict the saliency scores, it remains highly coupled with the encoder.

Different from those methods, we introduce CNN structures in the decoder. CNNs can capture local details, ensuring accurate saliency predictions across varying queries within the same video. The structure of the lightweight decoder for HD prediction is illustrated in \cref{fig:feedbacks}, which alternates between \textit{Linear} and \textit{Conv1d} layers. We denote the input of the HD decoder as $\mathcal{J} \in \mathbb{R} ^{N\times d}$. The saliency prediction $Y_\mathrm{hd} \in \mathbb{R} ^{N\times 1}$ can be computed as follows:
\begin{equation}
\label{eq:saliency_score}
    \begin{split}   
    Y_\mathrm{hd} =D_{\mathrm{HD}}\left( \mathcal{J} \right) +\frac{\mathbb{I} _{i=1}^{N}\left( \sum\nolimits_{k=1}^d{\mathcal{J} ^{ik}} \right)}{\sqrt{d}},
    \end{split}    
\end{equation}
where $\mathbb{I} _{i=1}^{N}\left( \cdot \right)$ represents the traversal operation. $D_{\mathrm{HD}}\left( \cdot \right)$ refers to the decoder network operation.
% Although QD-DETR \cite{moon2023qd-detr} designed the global saliency token for the HD prediction, this approach is highly coupled with the encoder. 

\textbf{Moment-guided Feedback.} We propose a moment-guided feedback manner to investigate the influence of MR on HD. Our key focus is on how to effectively utilize the results of moment retrieval for highlight detection. The implementation procedure of moment-guided feedback is illustrated in \cref{fig:feedbacks} (a).

We transform the tentatively predicted moment results from the ``(center, width)'' format to the ``(start, end)'' format. These moment results are fed into the Moment-to-Mask (M2M) converter to generate moment-aware masks $\xi _\mathrm{m2m}$. We initialize a clip-wise vector with a length of $N$ for each moment prediction. The indices in these vectors correspond to the clip indexes within the video. We binarize these vectors based on the moment results. We can obtain the clip indexes that are included in the moments through the prior information and set the values at those positions in the vector to 1, while the rest to 0. The obtained $N_q$ binary vectors are then summed and normalized with L2 norm.

The obtained moment-aware masks $\xi _\mathrm{m2m}$ integrate the prediction results from $N_q$ queries and provide guidance for the highlight detection. In the moment-guided feedback manner, the input $\mathcal{J}$ of the HD decoder is updated by $X_{\mathrm{hd}}+X_{\mathrm{hd}}*\xi _\mathrm{m2m}$.

\textbf{Highlightness-guided Feedback.} Similar to the moment-guided feedback, we introduce a highlightness-guided feedback manner with the Highlightness-to-Mask (H2M) converter, to explore the influence of HD on MR. As illustrated in \cref{fig:feedbacks} (b), the saliency score vector is binarized in the H2M to obtain the highlightness-aware mask $\xi _\mathrm{h2m}$. Values in the saliency score vector that are below the mean of the vector are set to 0, while those above are set to 1. Then $\xi _\mathrm{h2m}$ would provide guidance for moment retrieval. In this feedback manner, the input of the MR decoder is updated by $X_{\mathrm{mr}}+X_{\mathrm{mr}}*\xi _\mathrm{h2m}$.

\subsection{Task-dependent Joint Loss}
A direct way to optimize the joint task is to sum the respective task-specific loss functions. However, it neglects the variations in magnitudes of different task losses, leading to dominance of one task. Existing methods \cite{lei2021moment-detr,liu2022umt,moon2023qd-detr,jang2023knowing,lin2023univtg} utilize manually-set weights for the weighted sum of task losses, which limit the task learning as tasks evolve at different rates. To address this problem, inspired by the studies using dynamic weights \cite{chen2018gradnorm,liu2019end,guo2018dynamic,kendall2018multi,kendall2017uncertainties} based on task learning stages, we introduce the task-dependent joint loss as a more effective and flexible solution.

\textbf{MR Loss.} For moment retrieval, we define the MR likelihood using a Gaussian distribution as shown in \cref{eq:mr-likelihood}, with its mean determined by the output $f^{\theta _{\mathrm{mr}}}\left( x \right)$ of the neural network with $\theta _{\mathrm{mr}}$. $x$ is the input of the neural network.
\begin{equation}
\label{eq:mr-likelihood}  
    p\left( Y_{\mathrm{mr}}|f^{\theta _{\mathrm{mr}}}\left( x \right) \right) =\mathcal{N} \left( f^{\theta _{\mathrm{mr}}}\left( x \right) ,\delta _{\mathrm{mr}} \right),  
\end{equation}
where $\delta _{\mathrm{mr}}$ is a learnable parameter that quantifies the uncertainty of moment retrieval. The negative log-likelihood can be derived as follows:
\begin{equation}
\label{eq:mr-log-likelihood} 
    \begin{aligned}
    &-\log p\left(Y_{\mathrm{mr}}|f^{\theta _{\mathrm{mr}}}\left( x \right) \right) \\
    &\,\,\,\,\,\,\,\,\,\,\,\,\,\,\,\,\,\,\,\,\,\,\,\, \propto \frac{1}{2{\delta _{\mathrm{mr}}}^2}\left\| Y_{\mathrm{mr}}-f^{\theta _{\mathrm{mr}}}\left( x \right) \right\| ^2+\log \delta _{\mathrm{mr}}. 
    \end{aligned}
\end{equation}

$\left\| Y_{\mathrm{mr}}-f^{\theta _{\mathrm{mr}}}\left( x \right) \right\| ^2$ measures the offset between the model's predicted value and the ground-truth value. Therefore, inspired by \cref{eq:mr-log-likelihood}, the MR loss function $\mathcal{L} _{\mathrm{mr}}\left( \theta _{\mathrm{mr}},\delta _{\mathrm{mr}} \right) $ can be defined as:
\begin{equation}
\label{eq:mr-loss} 
    \begin{aligned}
    \mathcal{L} _{\mathrm{mr}} & \left( \theta _{\mathrm{mr}},\delta _{\mathrm{mr}}\right) = \frac{1}{2{\delta _{\mathrm{mr}}}^2}\mathcal{L} \left( \theta _{\mathrm{mr}} \right) +\log \delta _{\mathrm{mr}}.
    \end{aligned}
\end{equation}

Following existing approaches \cite{lei2021moment-detr,liu2022umt,moon2023qd-detr}, $\mathcal{L} \left( \theta _{\mathrm{mr}} \right) $ consists of three components: the L1 loss $L_{\mathrm{L1}}$, the generalized IoU loss \cite{rezatofighi2019giou} $L_{\mathrm{gIoU}}$, and the cross-entropy loss $L_{\mathrm{BCE}}$. $L_{\mathrm{L1}}$ and $L_{\mathrm{gIoU}}$ are employed to calculate the mean absolute error and gIoU deviation between ground-truth moments and predicted moments, respectively. $L_{\mathrm{BCE}}$ is used to classify whether the predicted moments belong to the foreground or background. In summary, $\mathcal{L} \left( \theta _{\mathrm{mr}} \right)=L_{\mathrm{L1}}+L_{\mathrm{gIoU}}+L_{\mathrm{BCE}}$.

% TVSum & Charades-STA
\begin{table*}
    \centering
    \begin{minipage}[b]{0.55\textwidth}
        \centering
        \renewcommand{\arraystretch}{1.3}%调行距
        \setlength\tabcolsep{1.2pt}%调列距
        \scalebox{0.9}{
        \begin{tabular}[b]{lccccccccccc}
    \toprule
    % ---------------------------------- table head ----------------------------------
      \multicolumn{1}{c|}{Method}  & \multicolumn{1}{c}{VT} &\multicolumn{1}{c}{VU} &\multicolumn{1}{c}{GA} 
      & \multicolumn{1}{c}{MS} &\multicolumn{1}{c}{PK} &\multicolumn{1}{c}{PR} &\multicolumn{1}{c}{FM} 
      & \multicolumn{1}{c}{BK} &\multicolumn{1}{c}{BT} &\multicolumn{1}{c}{DS} &\multicolumn{1}{c}{\textbf{Avg.}} \\
      \midrule
      
        \multicolumn{1}{l|}{sLSTM \cite{zhang2016slstm}\scriptsize{\textit{ECCV'16}}} & \multicolumn{1}{c}{41.1} &\multicolumn{1}{c}{46.2} &\multicolumn{1}{c}{46.3} &\multicolumn{1}{c}{47.7} &\multicolumn{1}{c}{44.8} &\multicolumn{1}{c}{46.1} &\multicolumn{1}{c}{45.2} &\multicolumn{1}{c}{40.6} &\multicolumn{1}{c}{47.1} &\multicolumn{1}{c}{45.5} &\multicolumn{1}{c}{45.1}\\
        
        \multicolumn{1}{l|}{SG \cite{mahasseni2017sg}\scriptsize{\textit{CVPR'17}}} & \multicolumn{1}{c}{42.3} &\multicolumn{1}{c}{47.2} &\multicolumn{1}{c}{47.5} &\multicolumn{1}{c}{48.9} &\multicolumn{1}{c}{45.6} &\multicolumn{1}{c}{47.3} &\multicolumn{1}{c}{46.4} &\multicolumn{1}{c}{41.7} &\multicolumn{1}{c}{48.3} &\multicolumn{1}{c}{46.6} &\multicolumn{1}{c}{46.2}\\
                    
        \multicolumn{1}{l|}{LIM-S \cite{xiong2019lims}\scriptsize{\textit{CVPR'19}}} & \multicolumn{1}{c}{55.9} &\multicolumn{1}{c}{42.9} &\multicolumn{1}{c}{61.2} &\multicolumn{1}{c}{54.0} &\multicolumn{1}{c}{60.4} &\multicolumn{1}{c}{47.5} &\multicolumn{1}{c}{43.2} &\multicolumn{1}{c}{66.3} &\multicolumn{1}{c}{69.1} &\multicolumn{1}{c}{62.6} &\multicolumn{1}{c}{56.3 }\\
        
        \multicolumn{1}{l|}{Trailer \cite{wang2020trailer}\scriptsize{\textit{ECCV'20}}} & \multicolumn{1}{c}{61.3} &\multicolumn{1}{c}{54.6} &\multicolumn{1}{c}{65.7} &\multicolumn{1}{c}{60.8} &\multicolumn{1}{c}{59.1} &\multicolumn{1}{c}{70.1} &\multicolumn{1}{c}{58.2} &\multicolumn{1}{c}{64.7} &\multicolumn{1}{c}{65.6} &\multicolumn{1}{c}{68.1} &\multicolumn{1}{c}{62.8}\\
        
        \multicolumn{1}{l|}{SL-Module \cite{xu2021slmodule}\scriptsize{\textit{ICCV'21}}} & \multicolumn{1}{c}{86.5} &\multicolumn{1}{c}{68.7} &\multicolumn{1}{c}{74.9} &\multicolumn{1}{c}{86.2} &\multicolumn{1}{c}{79.0} &\multicolumn{1}{c}{63.2} &\multicolumn{1}{c}{58.9} &\multicolumn{1}{c}{72.6} &\multicolumn{1}{c}{78.9} &\multicolumn{1}{c}{64.0} &\multicolumn{1}{c}{73.3}\\
        
        \multicolumn{1}{l|}{MINI-Net$^\dagger$ \cite{hong2020mininet}\scriptsize{\textit{ECCV'20}}} & \multicolumn{1}{c}{80.6} &\multicolumn{1}{c}{68.3} &\multicolumn{1}{c}{78.2} &\multicolumn{1}{c}{81.8} &\multicolumn{1}{c}{78.1} &\multicolumn{1}{c}{65.8} &\multicolumn{1}{c}{57.8} &\multicolumn{1}{c}{75.0} &\multicolumn{1}{c}{80.2} &\multicolumn{1}{c}{65.5} &\multicolumn{1}{c}{73.2}\\
        
        \multicolumn{1}{l|}{TCG$^\dagger$ \cite{ye2021tcg}\scriptsize{\textit{ICCV'21}}} & \multicolumn{1}{c}{85.0} &\multicolumn{1}{c}{71.4} &\multicolumn{1}{c}{81.9} &\multicolumn{1}{c}{78.6} &\multicolumn{1}{c}{80.2} &\multicolumn{1}{c}{75.5} &\multicolumn{1}{c}{71.6} &\multicolumn{1}{c}{77.3} &\multicolumn{1}{c}{78.6} &\multicolumn{1}{c}{68.1} &\multicolumn{1}{c}{76.8}\\
        
        \multicolumn{1}{l|}{Joint-VA$^\dagger$ \cite{badamdorj2021jointva}\scriptsize{\textit{ICCV'21}}} & \multicolumn{1}{c}{83.7} &\multicolumn{1}{c}{57.3} &\multicolumn{1}{c}{78.5} &\multicolumn{1}{c}{86.1} &\multicolumn{1}{c}{80.1} &\multicolumn{1}{c}{69.2} &\multicolumn{1}{c}{70.0} &\multicolumn{1}{c}{73.0} &\multicolumn{1}{c}{97.4} &\multicolumn{1}{c}{67.5} &\multicolumn{1}{c}{76.3 }\\
        \midrule
        
        \rowcolor{gray!10}
        \multicolumn{1}{l|}{UMT$^\dagger$ \cite{liu2022umt}\scriptsize{\textit{CVPR'22}}} & \multicolumn{1}{c}{87.5} &\multicolumn{1}{c}{81.5} &\multicolumn{1}{c}{88.2} &\multicolumn{1}{c}{78.8} &\multicolumn{1}{c}{81.4} &\multicolumn{1}{c}{\textbf{87.0}} &\multicolumn{1}{c}{76.0} &\multicolumn{1}{c}{86.9} &\multicolumn{1}{c}{84.4} &\multicolumn{1}{c}{79.6} &\multicolumn{1}{c}{83.1}\\

        \rowcolor{gray!10}
        \multicolumn{1}{l|}{QD-DETR \cite{moon2023qd-detr}\scriptsize{\textit{CVPR'23}}} & \multicolumn{1}{c}{88.2} &\multicolumn{1}{c}{87.4} &\multicolumn{1}{c}{85.6} &\multicolumn{1}{l}{85.0} &\multicolumn{1}{c}{85.8} &\multicolumn{1}{c}{86.9} &\multicolumn{1}{c}{76.4} &\multicolumn{1}{c}{91.3} &\multicolumn{1}{c}{89.2} &\multicolumn{1}{c}{73.7} &\multicolumn{1}{c}{85.0}\\       

        \rowcolor{gray!10}
        \multicolumn{1}{l|}{UniVTG$^\ddagger$ \cite{lin2023univtg}\scriptsize{\textit{ICCV'23}}} & \multicolumn{1}{c}{83.9} &\multicolumn{1}{c}{85.1} &\multicolumn{1}{c}{89.0} &\multicolumn{1}{l}{80.1} &\multicolumn{1}{c}{84.6} &\multicolumn{1}{c}{81.4} &\multicolumn{1}{c}{70.9} &\multicolumn{1}{c}{91.7} &\multicolumn{1}{c}{73.5} &\multicolumn{1}{c}{69.3} &\multicolumn{1}{c}{81.0}\\   

        \rowcolor{gray!10}
        \multicolumn{1}{l|}{\textbf{TaskWeave(Ours)}} & \multicolumn{1}{c}{\textbf{88.2}} &\multicolumn{1}{c}{\textbf{90.8}} &\multicolumn{1}{c}{\textbf{93.3}} &\multicolumn{1}{c}{\textbf{87.5}} &\multicolumn{1}{c}{\textbf{87.0}} &\multicolumn{1}{c}{82.0} &\multicolumn{1}{c}{\textbf{80.9}} &\multicolumn{1}{c}{\textbf{92.9}} &\multicolumn{1}{c}{\textbf{89.5}} &\multicolumn{1}{c}{\textbf{81.2}} &\multicolumn{1}{c}{\textbf{87.3}}\\ 
        \bottomrule
        \end{tabular}
        }
        \caption{Experimental results (\%) on TVSum. $\dagger$ means including audio modality. $\ddagger$ means following the pretrain-finetune paradigm.\\}
        \label{table:TVSum-res}
    \end{minipage}
    \hfill
    \begin{minipage}[b]{0.4\textwidth}
        \centering
        \renewcommand{\arraystretch}{1.3}%调行距
        \setlength\tabcolsep{1.2pt}%调列距
        \scalebox{0.9}{
        \begin{tabular}{c|ccc}
        \toprule
        % ---------------------------------- table head ----------------------------------
        \multicolumn{1}{c|}{\multirow{1}{*}{Backbone}} & 
        \multicolumn{1}{c}{\multirow{1}{*}{Method}}&
        \multicolumn{1}{c}{\multirow{1}{*}{R1@0.5}}&
        \multicolumn{1}{c}{\multirow{1}{*}{R1@0.7}}\\
        \midrule
        
        \multirow{7}{*}{\makecell*[c]{VGG}}& \multicolumn{1}{l}{SAP \cite{chen2019sap}\scriptsize{\textit{AAAI'19}}} &\multicolumn{1}{c}{27.42} & \multicolumn{1}{c}{13.36} \\
    
        {}&\multicolumn{1}{l}{SM-RL \cite{wang2019smrl}\scriptsize{\textit{CVPR'19}}}  &\multicolumn{1}{c}{24.36} & \multicolumn{1}{c}{11.17} \\
    
        {}&\multicolumn{1}{l}{2D-TAN \cite{zhang2020learning}\scriptsize{\textit{AAAI'20}}}  &\multicolumn{1}{c}{40.94} & \multicolumn{1}{c}{22.85} \\
    
        {}&\multicolumn{1}{l}{FVMR \cite{gao2021fvmr}\scriptsize{\textit{CVPR'21}}}  &\multicolumn{1}{c}{24.36} & \multicolumn{1}{c}{11.17} \\
        
        % \rowcolor{gray!20}
        {}& \multicolumn{1}{>{\columncolor{gray!10}}l}{UMT$^\dagger$ \cite{liu2022umt}\scriptsize{\textit{CVPR'22}}}  &\multicolumn{1}{>{\columncolor{gray!10}}c}{48.31} & \multicolumn{1}{>{\columncolor{gray!10}}c}{29.25} \\
    
        {}&\multicolumn{1}{>{\columncolor{gray!10}}l}{QD-DETR \cite{moon2023qd-detr}\scriptsize{\textit{CVPR'23}}}  &\multicolumn{1}{>{\columncolor{gray!10}}c}{52.77} & \multicolumn{1}{>{\columncolor{gray!10}}c}{31.13} \\
    
        {}&\multicolumn{1}{>{\columncolor{gray!10}}l}{\textbf{TaskWeave(Ours)}}  &\multicolumn{1}{>{\columncolor{gray!10}}c}{\textbf{56.51}} & \multicolumn{1}{>{\columncolor{gray!10}}c}{\textbf{33.66}} \\
        
        \midrule
        
        \multirow{5}{*}{\makecell*[c]{I3D}} &\multicolumn{1}{l}{CTRL \cite{gao2017tall}\scriptsize{\textit{ICCV'17}}} &\multicolumn{1}{c}{23.63} & \multicolumn{1}{c}{8.89}\\
    
        {}&\multicolumn{1}{l}{MAN \cite{zhang2019man}\scriptsize{\textit{CVPR'19}}} &\multicolumn{1}{c}{46.53} & \multicolumn{1}{c}{22.72}\\
        
        {}&\multicolumn{1}{l}{VSLNet \cite{zhang2020span}\scriptsize{\textit{ACL'20}}}  &\multicolumn{1}{c}{47.31} & \multicolumn{1}{c}{30.19}\\

        {} &\multicolumn{1}{>{\columncolor{gray!10}}l}{QD-DETR \cite{moon2023qd-detr}\scriptsize{\textit{CVPR'23}}} &\multicolumn{1}{>{\columncolor{gray!10}}c}{50.67} & \multicolumn{1}{>{\columncolor{gray!10}}c}{31.02}\\
        
        {}&\multicolumn{1}{>{\columncolor{gray!10}}l}{\textbf{TaskWeave(Ours)}}  &\multicolumn{1}{>{\columncolor{gray!10}}c}{\textbf{53.36}} & \multicolumn{1}{>{\columncolor{gray!10}}c}{\textbf{31.4}}\\ 
          
        \bottomrule
        \end{tabular}
        }
        \caption{Experimental results (\%) on Charades-STA test split. $\dagger$ means including audio modality. $\ddagger$ means following the pretrain-finetune paradigm.}
        \label{table:Charades-res}
    \end{minipage}
\end{table*}

% QVHighlights & component-ablation res
\begin{table*}[t]\normalsize
    \centering
    \begin{minipage}[b]{0.55\textwidth}
    \renewcommand{\arraystretch}{1.2}%调行距
    \setlength\tabcolsep{1.3pt}%调列距
    \scalebox{0.9}{
    \begin{tabular}{lccccccc}
    \toprule
    % ---------------------------------- table head ----------------------------------
    \multicolumn{1}{c}{\multirow{3}{*}{\textbf{Method}}} & 
    % \multicolumn{1}{c}{\multirow{3}{*}{Venue}} & 
      \multicolumn{5}{c}{\textbf{MR}} & \multicolumn{2}{c}{\textbf{HD}}\\ \cmidrule(lr){2-6} \cmidrule(lr){7-8}
            {}  &\multicolumn{2}{c}{R1}& \multicolumn{3}{c}{mAP}
            & \multicolumn{2}{c}{$\geq$ \text{Very Good}} \\
            \cmidrule(lr){2-3} \cmidrule(lr){4-6} \cmidrule(lr){7-8}
            {}  & \multicolumn{1}{c}{@0.5} &\multicolumn{1}{c}{@0.7} \quad &\multicolumn{1}{c}{@0.5} &\multicolumn{1}{c}{@0.75} &\multicolumn{1}{c}{Avg.} &\multicolumn{1}{c}{mAP} &\multicolumn{1}{c}{HIT@1}\\
            
    \midrule
    Moment-DETR \cite{lei2021moment-detr}\scriptsize{\textit{NIPS'21}}&\multicolumn{1}{c}{53.94} & \multicolumn{1}{c}{34.84} &\multicolumn{1}{c}{-} &\multicolumn{1}{c}{-} &\multicolumn{1}{c}{32.20} &\multicolumn{1}{c}{35.65} &\multicolumn{1}{c}{55.55} \\ 

     UMT$^\dagger$ \cite{liu2022umt}\scriptsize{\textit{CVPR'22}}&\multicolumn{1}{c}{60.26} & \multicolumn{1}{c}{44.26} &\multicolumn{1}{c}{-} &\multicolumn{1}{c}{-} &\multicolumn{1}{c}{38.59} &\multicolumn{1}{c}{\textbf{39.85}} &\multicolumn{1}{c}{\textbf{64.19}} \\ 

    QD-DETR \cite{moon2023qd-detr}\scriptsize{\textit{CVPR'23}}&\multicolumn{1}{c}{62.68} & \multicolumn{1}{c}{46.66} &\multicolumn{1}{c}{62.23} &\multicolumn{1}{c}{41.82} &\multicolumn{1}{c}{41.22} &\multicolumn{1}{c}{39.13} &\multicolumn{1}{c}{63.03} \\ 
     
    EaTR \cite{jang2023knowing}\scriptsize{\textit{ICCV'23}}&\multicolumn{1}{c}{61.36} & \multicolumn{1}{c}{45.79} &\multicolumn{1}{c}{61.86} &\multicolumn{1}{c}{41.91} &\multicolumn{1}{c}{41.74} &\multicolumn{1}{c}{37.15} &\multicolumn{1}{c}{58.65} \\ 

    UniVTG$^\ddagger$ \cite{lin2023univtg}\scriptsize{\textit{ICCV'23}}&\multicolumn{1}{c}{59.74} & \multicolumn{1}{c}{-} &\multicolumn{1}{c}{-} &\multicolumn{1}{c}{-} &\multicolumn{1}{c}{36.13} &\multicolumn{1}{c}{38.83} &\multicolumn{1}{c}{61.81} \\ 

    \rowcolor{gray!10}
    \textbf{TaskWeave(Ours)}&\multicolumn{1}{c}{\textbf{64.26}} & \multicolumn{1}{c}{\textbf{50.06}} &\multicolumn{1}{c}{\textbf{65.39}} &\multicolumn{1}{c}{\textbf{46.47}} &\multicolumn{1}{c}{\textbf{45.38}} &\multicolumn{1}{c}{39.28} &\multicolumn{1}{c}{63.68} \\ 
    
    \bottomrule
    \end{tabular}
    }
    \caption{Experimental results (\%) on QVHighlights val split. $\dagger$ means including audio modality. $\ddagger$ means following the pretrain-finetune paradigm.}
    \label{tab:QVHighlights-res}
    \end{minipage}
    \hfill
    \begin{minipage}[b]{0.4\textwidth}
        \centering
        \renewcommand{\arraystretch}{1.2}%调行距
        \setlength\tabcolsep{1.3pt}%调列距
        \scalebox{0.9}{
       \begin{tabular}{ccccccc}
    \toprule
    % ---------------------------------- table head ----------------------------------
    \multicolumn{1}{c}{\multirow{4}{*}{\makecell*[c]{Task-\\decoupled\\Unit}}} & 
    \multicolumn{1}{c}{\multirow{4}{*}{\makecell*[c]{Inter-\\task\\feedback}}} &
    \multicolumn{1}{c}{\multirow{4}{*}{\makecell*[c]{Joint\\loss}}} &
    \multicolumn{2}{c}{\textbf{MR}} & \multicolumn{2}{c}{\textbf{HD}}\\ 
      \cmidrule(lr){4-5} \cmidrule(lr){6-7}
        {}& {}& {} &\multicolumn{1}{c}{\makecell*[c]{R1\\@0.7}}& \multicolumn{1}{c}{\makecell*[c]{Avg.\\mAP}} & \multicolumn{1}{c}{mAP} &\multicolumn{1}{c}{HIT@1}\\
            
    \midrule
    {}& {}& {}& \multicolumn{1}{c}{46.26} & \multicolumn{1}{c}{41.0} &\multicolumn{1}{c}{38.94} &\multicolumn{1}{c}{62.84} \\ 
    
    {$\checkmark$}& {}& {}& \multicolumn{1}{c}{47.87} & \multicolumn{1}{c}{43.24} &\multicolumn{1}{c}{38.58} &\multicolumn{1}{c}{61.81} \\ 
    
    {$\checkmark$}& {}& {$\checkmark$}& \multicolumn{1}{c}{49.29} & \multicolumn{1}{c}{45.12} &\multicolumn{1}{c}{38.96} &\multicolumn{1}{c}{62.0} \\

    \rowcolor{gray!10}
    {$\checkmark$}& {$\checkmark$}& {$\checkmark$}& \multicolumn{1}{c}{50.06} & \multicolumn{1}{c}{45.38} &\multicolumn{1}{c}{39.28} &\multicolumn{1}{c}{63.68} \\ 
    
    \bottomrule
    \end{tabular}
    }    
    \caption{The ablation results (\%) of the components of our proposed method.}
    \label{table:components-ablation}
    \end{minipage}
\end{table*}

% inspired by the Boltzmann distribution, 
\textbf{HD Loss.} For highlight detection, inspired by the Boltzmann distribution, the HD likelihood is modeled using a softmax function applied to the scaled model output similar to \cite{kendall2018multi}, as \cref{eq:hd-likelihood}. The learnable scaling factor is denoted as $\delta _{\mathrm{hd}}$. $f^{\theta _{\mathrm{hd}}}\left( x \right) $ is the output of the network with $\theta _{\mathrm{hd}}$ on input $x$.
\begin{equation}
\label{eq:hd-likelihood}
    \begin{split}   
    p\left( Y_{\mathrm{hd}}|f^{\theta _{\mathrm{hd}}}\left( x \right) ,\delta _{\mathrm{hd}} \right) = \text{softmax} \left( \frac{1}{{\delta _{\mathrm{hd}}}^2}f^{\theta _{\mathrm{hd}}}\left( x \right) \right).
    \end{split}    
\end{equation}

% 篇幅所限，先注释掉
% The log-likelihood can be calculated by as follows:
% \begin{equation}
% \label{eq:hd-log-likelihood}
%     % \begin{split}   
%     \begin{aligned}
%     &\log p\left( Y_{\mathrm{hd}}=y|f^{\theta _{\mathrm{hd}}}\left( x \right) ,\delta _{\mathrm{hd}} \right) \\
%     &\,\,\,\,\,\,\,\,\, =\frac{1}{{\delta _{\mathrm{hd}}}^2}f_{y}^{\theta _{\mathrm{hd}}}\left( x \right) -\log \sum_{y\prime}{\exp \left( \frac{1}{{\delta _{\mathrm{hd}}}^2}f_{y\prime}^{\theta _{\mathrm{hd}}}\left( x \right) \right)},
%     % \end{split}    
%     \end{aligned}
% \end{equation}
% where $f_{y}^{\theta _{\mathrm{hd}}}\left( x \right) $ refers to the y-th value of the $f^{\theta _{\mathrm{hd}}}\left( x \right) $. 

% HD loss function $\mathcal{L} _{\mathrm{hd}}\left( \theta _{\mathrm{hd}},\delta _{\mathrm{hd}} \right) $ can be calculated by: 
% & \mathcal{L} _{\mathrm{hd}}\left( \theta _{\mathrm{hd}},\delta _{\mathrm{hd}} \right) =

Similar to \cref{eq:mr-log-likelihood}, the negative log-likelihood of \cref{eq:hd-likelihood} can be calculated as follows:
\begin{equation}
\label{eq:hd-log-likelihood} 
    \begin{aligned}
    &-\log p\left( Y_{\mathrm{hd}}=y|f^{\theta _{\mathrm{hd}}}\left( x \right) ,\delta _{\mathrm{hd}} \right)
    \\
    &\,\,\,\,\,\,\,\,\,\,\,\,\,\,=\frac{1}{{\delta _{\mathrm{hd}}}^2}\log \frac{\sum_{y\prime}{\exp \left( f_{y\prime}^{\theta _{\mathrm{hd}}}\left( x \right) \right)}}{\exp \left( f_{y}^{\theta _{\mathrm{hd}}}\left( x \right) \right)} 
    \\&\,\,\,\,\,\,\,\,\,\,\,\,\,\, +\log \frac{\sum_{y\prime}{\exp \left( \frac{1}{{\delta _{\mathrm{hd}}}^2}f_{y\prime}^{\theta _{\mathrm{hd}}}\left( x \right) \right)}}{\left( \sum_{y\prime}{\exp \left( f_{y\prime}^{\theta _{\mathrm{hd}}}\left( x \right) \right)} \right) ^{\frac{1}{{\delta _{\mathrm{hd}}}^2}}}
    \\
    &\,\,\,\,\,\,\,\,\,\,\,\,\,\,\approx -\frac{1}{{\delta _{\mathrm{hd}}}^2}\log \text{softmax} \left( Y_{\mathrm{hd}},f^{\theta _{\mathrm{hd}}}\left( x \right) \right) +\log \delta _{\mathrm{hd}},
    % &\,\,\,\,\,\,\,\,\,\,\,\,\,\,\approx \frac{1}{{\delta _{\mathrm{hd}}}^2}\mathcal{L}\left( \theta _{\mathrm{hd}} \right) +\log \delta _{\mathrm{hd}},
    \end{aligned}
\end{equation}
where $f_{y}^{\theta _{\mathrm{hd}}}\left( x \right) $ refers to the y-th value of the $f^{\theta _{\mathrm{hd}}}\left( x \right) $. We introduce a simplify assumption $\sum_{y\prime}{\exp \left( \frac{1}{{\delta _{\mathrm{hd}}}^2}f_{y\prime}^{\theta _{\mathrm{hd}}}\left( x \right) \right)} \approx \left( \sum_{y\prime}{\exp \left( f_{y\prime}^{\theta _{\mathrm{hd}}}\left( x \right) \right)} \right) ^{\frac{1}{{\delta _{\mathrm{hd}}}^2}}$ when $\delta _{\mathrm{hd}} \rightarrow 1$. 

$-\log \text{softmax} \left( Y_{\mathrm{hd}},f^{\theta _{\mathrm{hd}}}\left( x \right) \right)$ of \cref{eq:hd-log-likelihood} represents the cross-entropy classification loss of $Y_{\mathrm{hd}}$. Inspired on this term, we generalize this expression to other classification losses. The HD loss function $\mathcal{L} _{\mathrm{hd}}\left( \theta _{\mathrm{hd}},\delta _{\mathrm{hd}} \right) $ can be formulated as: 
\begin{equation}
\label{eq:hd-loss} 
    \begin{aligned}
    \mathcal{L} _{\mathrm{hd}}\left( \theta _{\mathrm{hd}},\delta _{\mathrm{hd}} \right) =\frac{1}{{\delta _{\mathrm{hd}}}^2}\mathcal{L} \left( \theta _{\mathrm{hd}} \right) +\log \delta _{\mathrm{hd}}.
\end{aligned}
\end{equation}
Consistent with prior approaches \cite{lei2021moment-detr,moon2023qd-detr}, $\mathcal{L}\left( \theta _{\mathrm{hd}} \right)$ includes the hinge loss $L_{\mathrm{hinge}}$, negative video-query pairs loss $L_{\mathrm{neg}}$, and rank-aware contrastive loss \cite{hoffmann2022ranking} $L_{\mathrm{cont}}$, i.e.  $\mathcal{L}\left( \theta _{\mathrm{hd}} \right) = L_{\mathrm{hinge}} + L_{\mathrm{neg}} + L_{\mathrm{cont}}$. $L_{\mathrm{hinge}}$ is computed between two pairs of positive and negative clips, with its margin 0.2 to maintain consistency with \cite{lei2021moment-detr} for fairness. $L_{\mathrm{neg}}$ and $L_{\mathrm{cont}}$ from \cite{moon2023qd-detr} are used to reduce the saliency of negative pairs.

We can obtain final task-dependent joint loss $\mathcal{L}_\mathrm{joint}$ through integrating \cref{eq:mr-loss} and \cref{eq:hd-loss}:
\begin{equation}
\label{eq:joint-loss} 
\begin{aligned}
&\mathcal{L} _{joint}\,=\mathcal{L} _{\mathrm{mr}}\left( \theta _{\mathrm{mr}},\delta _{\mathrm{mr}} \right) +\mathcal{L} _{\mathrm{hd}}\left( \theta _{\mathrm{hd}},\delta _{\mathrm{hd}} \right) .
% \\
% &=\,\frac{1}{2{\delta _{\mathrm{mr}}}^2}\mathcal{L} \left( \theta _{\mathrm{mr}} \right) +\log \delta _{\mathrm{mr}}+\frac{1}{{\delta _{\mathrm{hd}}}^2}\mathcal{L} \left( \theta _{\mathrm{hd}} \right) +\log \delta _{\mathrm{hd}}.
\end{aligned}
\end{equation}

\section{Experiments}
\label{sec:experiments}

% Ablations on decoupled unit & loss & feedback.
\begin{table*}[t]\normalsize
    \begin{minipage}[b]{0.35\linewidth}
    \centering
    \renewcommand{\arraystretch}{1.2}%调行距
    \setlength\tabcolsep{1.pt}%调列距
    \scalebox{0.9}{
    \begin{tabular}{ccccccc}
    \toprule
    % ---------------------------------- table head ----------------------------------
    \multicolumn{1}{c}{\multirow{4}{*}{Index}} &
    \multicolumn{1}{c}{\multirow{4}{*}{\makecell*[c]{MR-\\expert}}} &
    \multicolumn{1}{c}{\multirow{4}{*}{\makecell*[c]{HD-\\expert}}} &
    \multicolumn{2}{c}{\textbf{MR}} & \multicolumn{2}{c}{\textbf{HD}}\\ 
      \cmidrule(lr){4-5} \cmidrule(lr){6-7}
        {}& {}& {} &\multicolumn{1}{c}{\makecell*[c]{R1\\@0.7}}& \multicolumn{1}{c}{\makecell*[c]{Avg.\\mAP}} & \multicolumn{1}{c}{mAP} &\multicolumn{1}{c}{HIT@1}\\
            
    \midrule
    {(a)}& {Iden.}& {Iden.}& \multicolumn{1}{c}{44.77} & \multicolumn{1}{c}{40.23} &\multicolumn{1}{c}{38.38} &\multicolumn{1}{c}{60.19} \\ 
    
    {(b)}&{Line.}& {Line.}& \multicolumn{1}{c}{47.55} & \multicolumn{1}{c}{43.65} &\multicolumn{1}{c}{38.32} &\multicolumn{1}{c}{60.77} \\ 

    {(c)}&{CNN}& {CNN}& \multicolumn{1}{c}{49.23} & \multicolumn{1}{c}{43.73} &\multicolumn{1}{c}{38.82} &\multicolumn{1}{c}{60.77} \\
        
    {(d)}&{Trans.}& {Trans.}& \multicolumn{1}{c}{48.32} & \multicolumn{1}{c}{42.85} &\multicolumn{1}{c}{38.61} &\multicolumn{1}{c}{61.10} \\

    {(e)}&{Trans.}& {Iden.}& \multicolumn{1}{c}{47.1} & \multicolumn{1}{c}{42.1} &\multicolumn{1}{c}{38.7} &\multicolumn{1}{c}{61.68} \\

    {(f)}&{CNN}& {Iden.}& \multicolumn{1}{c}{49.29} & \multicolumn{1}{c}{45.12} &\multicolumn{1}{c}{38.96} &\multicolumn{1}{c}{62.0} \\

    \bottomrule
    \end{tabular}
    }
    \caption{Flexibility validation (\%) of the task-decoupled unit. ``Iden.'': identity mapping; ``Trans.'': Transformer; ``Line.'': linear layer.}
    \label{tab:Ablations on decoupled unit.}
    \end{minipage}
    \hfill
    \begin{minipage}[b]{0.3\linewidth}
        \centering
        \renewcommand{\arraystretch}{1.2}%调行距
        \setlength\tabcolsep{1.pt}%调列距
        \scalebox{0.9}{
        \begin{tabular}{ccccc}
    \toprule
    % ---------------------------------- table head ----------------------------------
    \multicolumn{1}{c}{\multirow{4}{*}{Type}} &
    \multicolumn{2}{c}{\textbf{MR}} & \multicolumn{2}{c}{\textbf{HD}}\\ 
      \cmidrule(lr){2-3} \cmidrule(lr){4-5}
        {}&\multicolumn{1}{c}{\makecell*[c]{R1\\@0.7}}& \multicolumn{1}{c}{\makecell*[c]{Avg.\\mAP}} & \multicolumn{1}{c}{mAP} &\multicolumn{1}{c}{HIT@1}\\
            
    \midrule
    {Sum}& \multicolumn{1}{c}{47.68} & \multicolumn{1}{c}{44.27} &\multicolumn{1}{c}{38.7} &\multicolumn{1}{c}{60.32} \\ 
    
    {\makecell*[c]{Weighted Sum}}& \multicolumn{1}{c}{48.32} & \multicolumn{1}{c}{43.21} &\multicolumn{1}{c}{38.31} &\multicolumn{1}{c}{61.35} \\ 

    {Ours}& \multicolumn{1}{c}{50.06} & \multicolumn{1}{c}{45.38} &\multicolumn{1}{c}{39.28} &\multicolumn{1}{c}{63.68} \\

    \bottomrule
    \end{tabular}
    }    
    \caption{Effectiveness justification (\%) of the task-dependent joint loss.\\          
     }
    \label{tab:Ablations on losses.}
    \end{minipage}
    \hfill
    \begin{minipage}[b]{0.3\linewidth}
        \centering
        \renewcommand{\arraystretch}{1.2}%调行距
        \setlength\tabcolsep{1.6pt}%调列距
        \scalebox{0.9}{
        \begin{tabular}{ccccc}
    \toprule
    % ---------------------------------- table head ----------------------------------
    \multicolumn{1}{c}{\multirow{4}{*}{Type}} &
    \multicolumn{2}{c}{\textbf{MR}} & \multicolumn{2}{c}{\textbf{HD}}\\ 
      \cmidrule(lr){2-3} \cmidrule(lr){4-5}
        {}&\multicolumn{1}{c}{\makecell*[c]{R1\\@0.7}}& \multicolumn{1}{c}{\makecell*[c]{Avg.\\mAP}} & \multicolumn{1}{c}{mAP} &\multicolumn{1}{c}{HIT@1}\\
            
    \midrule
    {MR2HD}& \multicolumn{1}{c}{50.06} & \multicolumn{1}{c}{45.38} &\multicolumn{1}{c}{38.73} &\multicolumn{1}{c}{62.84} \\ 
    
    {HD2MR}& \multicolumn{1}{c}{48.65} & \multicolumn{1}{c}{44.78} &\multicolumn{1}{c}{39.28} &\multicolumn{1}{c}{63.68} \\ 

    {Bi-MRHD}& \multicolumn{1}{c}{48.84} & \multicolumn{1}{c}{45.1} &\multicolumn{1}{c}{38.36} &\multicolumn{1}{c}{61.81} \\

    {MR-HD}& \multicolumn{1}{c}{50.0} & \multicolumn{1}{c}{45.3} &\multicolumn{1}{c}{38.96} &\multicolumn{1}{c}{62.45} \\

    {HD-MR}& \multicolumn{1}{c}{49.1} & \multicolumn{1}{c}{44.87} &\multicolumn{1}{c}{39.03} &\multicolumn{1}{c}{59.94} \\
    \bottomrule
    \end{tabular}
    }    
    \caption{Comparing various inter-task feedback combinations (\%) on QVHighlights val split.}
    \label{tab:Ablations on feedback.}
    \end{minipage}
\end{table*}

\subsection{Experimental Setup}
\textbf{Datasets.} Extensive experiments are conducted on three benchmark datasets: QVHighlights \cite{lei2021moment-detr}, TVSum \cite{song2015tvsum}, and Charades-STA \cite{gao2017tall}. QVHighlights is currently the sole publicly dataset for joint moment retrieval and highlight detection tasks. It provides 10,310 queries associated with 18,367 moments, with an average of 1.8 disjoint moments per query. In contrast to other MR datasets with one-to-one query-moment mappings, QVHighlights aligns more closely with real-world scenarios. Each video in the dataset comprises 75 clips, each of which is 2s-long.

We also utilize two task-specific datasets, Charades-STA \cite{gao2017tall} for MR and TVSum \cite{song2015tvsum} for HD, to evaluate the model. Charades-STA provides 16,128 query-moment pairs. TVSum comprises videos from 10 domains, with each domain containing 5 videos.

\textbf{Evaluation Metrics.} We utilize the same evaluation metrics used in prior approaches \cite{moon2023qd-detr,liu2022umt,lei2021moment-detr}. For QVHighlights, mean average precision (mAP) with different tIoU thresholds {0.5, 0.75}, the average mAP over [0.5:0.05:0.95], and Recall@1 with tIoU {0.5, 0.7} are utilized for MR evaluation. mAP and HIT@1 are used for HD evaluation, where a clip is considered as a true positive when it achieves a ``Very Good'' \cite{lei2021moment-detr} saliency score. For Charades-STA, we utilize Recall@1 with tIoU thresholds {0.5, 0.7}. For TVSum, Top-5 mAP is adopted.

\subsection{Implementation Details}

\textbf{Feature representations.} For OVHighlights, we employ the pre-trained SlowFast \cite{feichtenhofer2019slowfast} and CLIP \cite{radford2021clip} backbones to extract video features, following \cite{lei2021moment-detr,moon2023qd-detr,liu2022umt} for fairness. For Charedes-STA, we leverage the official release of VGG \cite{simonyan2014vgg} and I3D \cite{carreira2017i3d} features as video embeddings. For TVSum, we extract video features by the I3D \cite{carreira2017i3d} pre-trained on Kinetics 400 \cite{kay2017kinetics400}. For QVHighlights and TVSum, we use CLIP \cite{radford2021clip} to extract text features, while using GloVe \cite{pennington2014glove} text embeddings for Charades-STA.

\textbf{Training settings.} We leverage AdamW \cite{kingma2014adam} optimizer with 1e-4 learning rate and 1e-4 weight decay. We train 200, 100, and 2000 epochs with batch size 32, 32 and 2 for QVHighlights, Charades-STA and TVSum, respectively. All Transformer layers follow the consistent configuration, including sinusoidal positional encodings, 8 attention heads, and a dropout rate of 0.1. Multi-modal fusion employs a 2-layer Transformer with cross attention, where video features serve as the \textit{query} and text features serve as \textit{key} and \textit{value}.  For \cref{eq:joint-loss}, we define $\gamma_\mathrm{mr},\gamma_\mathrm{hd}$ as $\log{\delta _{\mathrm{mr}}}^2,\log{\delta _{\mathrm{hd}}}^2$, respectively. Therefore, \cref{eq:joint-loss} can be rewritten as $\mathcal{L} _{joint}=\exp \left( -\gamma _{\mathrm{mr}} \right) \mathcal{L} \left( \theta _{\mathrm{mr}} \right) +2\exp \left( -\gamma _{\mathrm{hd}} \right) \mathcal{L} \left( \theta _{\mathrm{hd}} \right) +\gamma _{\mathrm{mr}}+\gamma _{\mathrm{hd}}$. $\gamma_\mathrm{mr}$ and $\gamma_\mathrm{hd}$ are learnable parameters. They are initialized to 0. To further stabilize the training, the model EMA strategy has also undertaken. All experiments are conducted with Pytorch v1.13.1 on a single NVIDIA RTX 3090.

\subsection{Comparison with State-of-the-arts}

\textbf{Results on TVSum.} In \cref{table:TVSum-res}, we compare our TaskWeave with existing state-of-the-arts (SOTA) methods, where the methods with $\dagger$ are incorporate with audio features. UniVTG \cite{lin2023univtg} follows pretrain-then-finetune paradigm, therefore we use its non-pretrained results for a fair comparison. We observe that: i) our methods outperforms SOTA methods in 9 out of the 10 categories; ii) TaskWeave significantly outperforms all methods in terms of Avg. mAP, with a remarkable 2.71\% improvement over the previous SOTA method.

\textbf{Results on Charades-STA.} In \cref{table:Charades-res}, we utilize different backbones to demonstrate the effectiveness of our TaskWeave. We compare our results with other methods for MR only (white background) and methods for joint MR and HD (gray background). It results in significant improvements, \textit{i.e.}, +7.09\% in R1@0.5 and +8.13\% in R1@0.7.

\textbf{Results on QVHighlights.} As shown in \cref{tab:QVHighlights-res}, we present the performance comparisons with existing methods. We report the MR results with the MR-guided feedback, while HD results with the HD-guided feedback. We observe that: i) our TaskWeave surpasses SOTA methods by a large margin in all moment retrieval metrics, with a remarkable 8.72\% improvement in Avg. mAP; ii) our method achieves the second-best performance on the HD. However, It's worth noting that UMT \cite{liu2022umt} incorporates audio features, while our method does not. When compared to other methods without audio features, our method outperforms them all. In summary, these results validate the effectiveness of our task-driven framework.

\begin{figure*}[htb]
  \centering
  \includegraphics[width=1.0\linewidth]{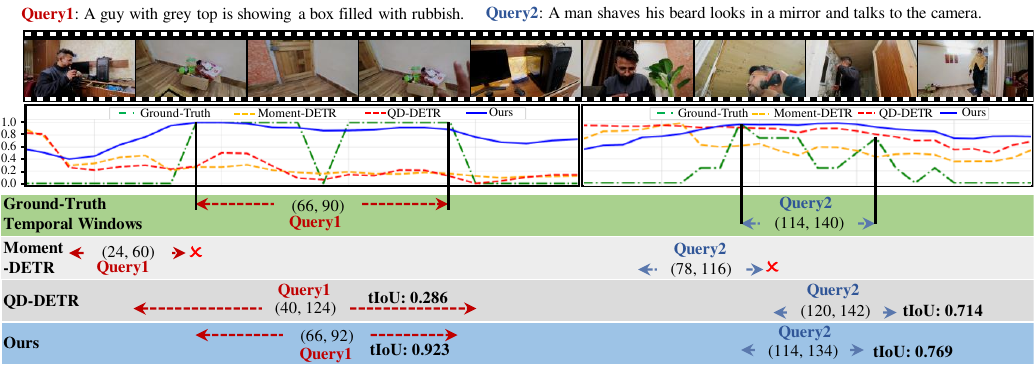}
  \caption{Qualitative results on the QVHighlights for Ground-Truth, Moment-DETR, QD-DETR and our method. The predicted moments and saliency scores are illustrated through intervals and lines. }
  \label{fig:comparison_visual}
\end{figure*}

We provide visualization examples in \cref{fig:comparison_visual} for the qualitative analysis among our TaskWeave, Moment-DETR \cite{lei2021moment-detr}, and QD-DETR \cite{moon2023qd-detr}. Given diverse queries for a video, TaskWeave precisely localizes moments for queries and presents high IoU with the ground-truth. Moment-DETR misses some instances, and QD-DETR exhibits lower retrieval accuracy. Moreover, our method obtains higher saliency scores for relevant clips in response to the query.

\subsection{Ablation Studies and Discussions}
We present some ablation studies and discussions about TaskWeave, with all experiments are conducted on the QVHighlights \textit{val} split \cite{lei2021moment-detr}.

\textbf{Ablation of components.} We conduct ablation studies on each component of TaskWeave, as shown in \cref{table:components-ablation}. Sequentially employing the task-decoupled unit and our proposed joint loss contributes a 5.5\% improvement in average mAP and a 6.5\% increase in R1@0.7 for MR. Utilizing all components, we observe that a significant 10.7\% improvement in average mAP of MR and a 1.3\% enhancement in HIT@1 of HD. These results indicate the effectiveness of the proposed components in out TaskWeave.

\textbf{Flexibility of the task-decoupled unit.}
To demonstrate the flexibility of the task-decoupled unit, we investigate the performance of applying different networks within various experts, as shown in \cref{tab:Ablations on decoupled unit.}. The shared expert utilize a fixed configuration with 2 Transformer layers. We provide 4 different methods for each task-specific expert, including the feed-forward network (one layer Linear), the identity mapping, CNN (composed of one depthwise convolution layer with the kernel/stride/padding of 5/1/2 and 1D convolution), and the Transformer. Due to space constraints, we present results of 6 out of 16 combinations. The results in \cref{tab:Ablations on decoupled unit.} are obtained by TaskWeave without inter-task feedback.

From this table, we can observe several interesting facts. First, different task-specific experts have different performance. The need for experts is reflected in (a). Second, although Transformer has made significant progress in visual, it's not a panacea. Results in (d) are lower compared to (b) and (c), we believe that the high computational complexity of Transformer decrease the performance. Third, the task-specific expert should be designed based on the task objective. For instance, (c) and (f) perform well on the MR because they focus on local features, which is important for localizing moments. Finally, we also find evidence that MR and HD are highly related. Improvement in one task enhances the other (compare (c) and (f)), while a decline in one limits the other's performance (compare (d) and (e)). In this paper, our mr-specific expert utilizes CNN and hd-specific expert is implemented by identity mapping.

\textbf{Task-dependent joint loss.} \cref{tab:Ablations on losses.} shows the performance of TaskWeave with different losses. ``Sum'' and ``Weighted Sum'' refer to the fusion manner of loss for different tasks, respectively. Weights in ``Weighted Sum'' are consistent with the existing methods \cite{lei2021moment-detr,moon2023qd-detr} for fairness. Comparing ``Sum'' and ``Weighted Sum'', we find that ``Weighted Sum'' can better balance the performance of the two tasks. However, it is obvious that the proposed task-dependent joint loss is optimal.

\textbf{Inter-task interactions.} We believe that each task requires learning before providing effective feedback. Therefore, in our inter-task feedback mechanism, the feedback process starts when the model is trained to half of the max epoch. For brevity, we write Moment/Highlightness-guided feedback as MgF/HgF. In \cref{tab:Ablations on feedback.}, we explore five feedback manners as follows: ``MR2HD'' (with MgF only), ``HD2MR'' (with HgF only), ``Bi-MRHD'' (with MgF and HgF simultaneously), ``MR-HD'' (MgF first, then HgF), and ``HD-MR'' (HgF first, then MgF).

We find that the inter-task feedback makes both tasks gain simultaneously. The performance of the model with ``HD2MR'' is slightly worse, this is because HD focuses on more refined moments than MR. Ground-truth annotations and the results of ``MR-HD'' and ``HD-MR'' also illustrate this fact. The gain brought by ``Bi-MRHD'' is limited, which we believe that it's a natural result of feedback not always being effective. In general, the inter-task feedback not only contributes to bring gains for both tasks but also helps to understand the characteristics of each task.

\section{Conclusion}
\label{sec:conclusion}

This paper proposes a novel task-driven paradigm for addressing joint moment retrieval and highlight detection. Different from existing data-driven methods, we utilize the task-decoupled unit to capture the task-specific and common features, respectively. We also explore different network architecture for moment retrieval and highlight detection. We design the inter-task feedback mechanism to in-depth investigate the interplay between both tasks. Different from prior methods, we introduce the principled joint loss to optimize the model. The effectiveness, flexibility, and superiority of the proposed method have been demonstrated on three benchmark datasets.

\section*{Acknowledgement}
This research was supported by the National Natural Science Foundation of China (No. U23B2060, 62088102), the Youth Innovation Team of Shaanxi Universities, and the Fundamental Research Funds for the Central Universities.
{
    \small
    \bibliographystyle{ieeenat_fullname}
    \bibliography{main}
}


\end{document}